\pdfoutput=1
\documentclass[11pt]{article}

\usepackage[preprint]{acl}

\usepackage{times}
\usepackage{latexsym}
\usepackage{amsmath}
\usepackage{booktabs}
\usepackage{amssymb}
\usepackage{adjustbox}
\usepackage{multirow} 
\usepackage[T1]{fontenc}

\usepackage[utf8]{inputenc}

\usepackage{microtype}

\usepackage{inconsolata}

\usepackage{graphicx}

\newcommand{\sys}{AnomalyJev}

\title{AnomalyJev: Towards Evaluating Image Anomaly Explanations \\ via Claim-Level Verification}

\author{
  \textbf{Qizhou Wang\textsuperscript{1}},
  \textbf{Jiawen Zhu\textsuperscript{2}},
  \textbf{Guansong Pang\textsuperscript{2}},
  \textbf{Christopher Leckie\textsuperscript{1}}
\\
  \textsuperscript{1}The University of Melbourne,
  \textsuperscript{2}Singapore Management University
}

\begin{document}
\maketitle
\begin{abstract}
Recent multimodal LLM (MLLM)-based anomaly detection (AD) methods can generate textual explanations that help interpret their predictions. 
However, reliably and efficiently evaluating these explanations remains challenging.
While LLM judges can assess explanations through structured rubrics, they often require substantial token generation, making evaluation costly and difficult to scale. 
We instead propose to anchor evaluation in the available ground truth by extracting verifiable claims from each explanation and verifying them against the corresponding annotations. 
We instantiate this idea as AnomalyJev, a lightweight framework that extracts verifiable claims from contextualised representations using ground-truth-defined queries and pointer based heads, without token generation. The claims are then verified against the ground truth to score faithfulness and completeness.
For evaluation, we construct a human-annotated set of 640 explanations with over 1,600 expert ratings across in-domain and cross-dataset settings. Despite its compact design, \sys{} achieves stronger agreement with human annotators than most open MLLM judges and approaches that of flagship MLLM judges, at a fraction of their runtime and resources.

\end{abstract}

\section{Introduction}
Image Anomaly Detection (AD) aims to identify images that deviate from normal or satisfactory conditions~\citep{ad_survey}, with applications ranging from industrial inspection to medical diagnosis~\citep{bergmann2019mvtec, roth2022patchcore}. Recent MLLM-based AD methods have achieved promising detection performance~\citep{gu2024anomalygpt, xu2025anomalyov}. Beyond anomaly detection, their ability to generate natural language has motivated growing interest in producing textual explanations to make anomaly predictions more interpretable~\citep{chao2025anomalyr1}. 

\begin{figure}
    \centering
    \includegraphics[width=\linewidth]{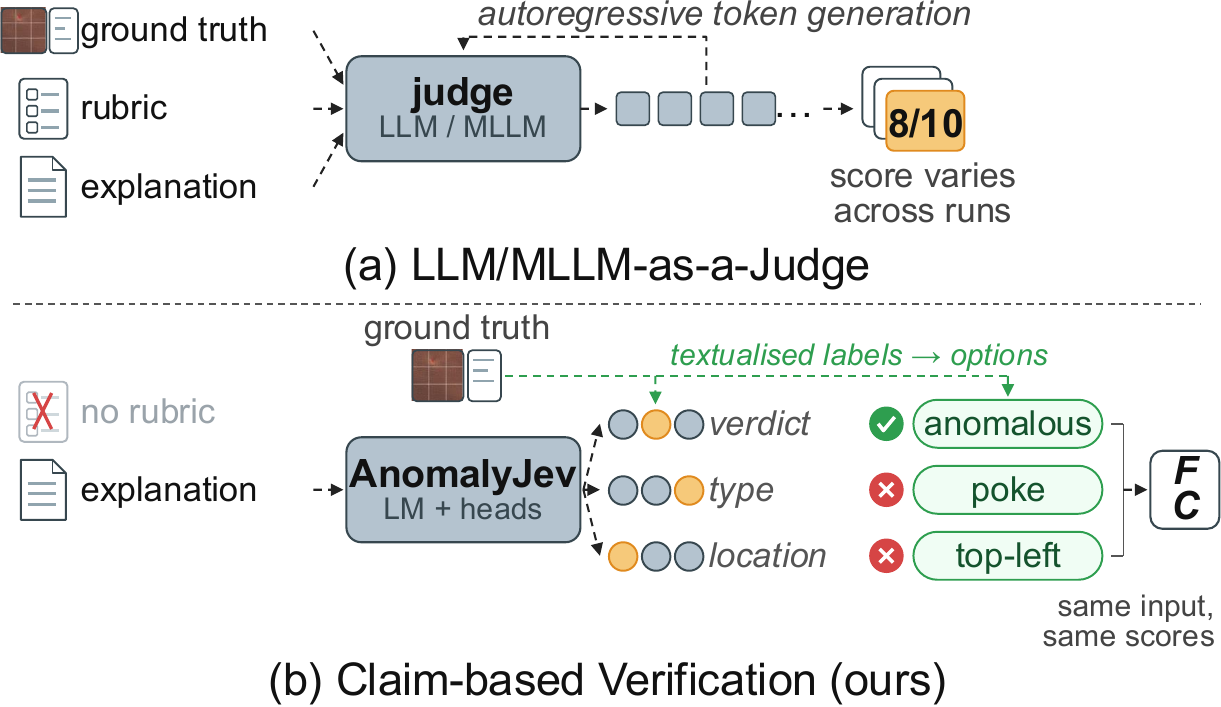}
\caption{LLM/MLLM-as-a-judge vs.\ claim-based verification.
(a) A judge takes the ground truth, evaluation rubric, and explanation as input and autoregressively generates a score.
(b) \sys{} predicts one option per ground-truth attribute directly from contextualised representations, then verifies the resulting claims to produce deterministic scores, without token generation.}
    \label{fig:intuition}
    \vspace{-0.5cm}
\end{figure}

However, evaluating image AD explanations remains challenging and underexplored. Existing MLLM-based AD methods often leave their generated explanations unevaluated or assess them using conventional metrics \citep{xu2025anomalyov, jiang2025mmad}, which may not adequately capture the anomaly-relevant factual details. Conventional metrics such as BLEU \citep{papineni2002bleu} struggle to assess anomaly-relevant details and their factual correctness. While MLLM judges can assess explanations using structured rubrics, they often incur substantial computational costs due to large model sizes and autoregressive token generation, limiting their scalability for repeated evaluation across models and datasets. Moreover, directly generated evaluation scores may vary across runs and can be influenced by linguistic fluency rather than factual correctness~\citep{zheng2023judging, ovad2026}.

To address these challenges, we propose to anchor evaluation in the available ground truth. An informative anomaly explanation should correctly describe key anomaly attributes, such as its type and location. We assess this along two complementary dimensions: \textit{faithfulness}, which measures the correctness of stated claims, and \textit{completeness}, which measures how much of the verifiable ground-truth information is correctly covered. As illustrated in Fig.~\ref{fig:intuition}, our approach extracts verifiable claims from free-text explanations and checks them against the corresponding ground truth annotations. By formulating claim extraction as selecting among candidate options for each attribute, we can perform this process efficiently using a small language model without autoregressive token generation.

We instantiate this idea in AnomalyJev, a lightweight framework named after Jev-style decision models that answer closed-set questions without autoregressive token generation. AnomalyJev formulates each verifiable attribute as a query with a set of candidate options and uses query-specific pointer heads~\citep{vinyals2015pointer} to select answers directly from the contextualised representations of a small LLM. The extracted claims are then verified against the available ground truth to compute faithfulness and completeness scores.
AnomalyJev is trained using weak labels generated by a commercial LLM, without requiring human annotations, in just a few GPU-hours. To evaluate its effectiveness, we collect 640 explanations from 11 MLLMs and obtain over 1,600 expert ratings. AnomalyJev achieves stronger agreement with human judgements than most open MLLM judges and approaches the performance of flagship proprietary judges, while running 40× faster at roughly 1/150 of the cost. Applying AnomalyJev to 11,586 explanations, with rankings confirmed by the human-rated subset, further reveals that anomaly-specialised MLLMs do not consistently produce better explanations than general-purpose MLLMs. 

Our contributions are summarised as follows:
\begin{itemize}
    \item We propose verifiable claim-based evaluation for MLLM-generated AD explanations, assessing the correctness of stated claims (faithfulness) and the coverage of verifiable ground-truth information (completeness). 
    \item We instantiate it with AnomalyJev, a lightweight evaluation framework that extracts and verifies verifiable claims without token generation, at a fraction of the cost of MLLM judges.
    \item We construct a human-annotated set of 640 explanations from 11 MLLMs with over 1,600 ratings from two experts, on which AnomalyJev shows competitive agreement with human judges on both faithfulness and completeness.
\end{itemize}

\section{Related Work}
\paragraph{Explanations for Image Anomaly Detection.}
Image anomaly detection has been extensively studied \citep{bergmann2019mvtec, roth2022patchcore}, while textual explanation is an emerging capability enabled by recent advances in MLLMs. Existing methods can be broadly divided into two categories. The first prompts general-purpose MLLMs with anomaly detection instructions and uses them as zero-shot or few-shot detectors \citep{cao2023gpt4v, jiang2025mmad}. The second develops dedicated MLLM-based models specifically for anomaly detection and explanation generation, from AnomalyGPT \citep{gu2024anomalygpt} and Myriad \citep{li2023myriad} to instruction-tuned and RL-tuned models such as Anomaly-OV \citep{xu2025anomalyov}, AnomalyR1 \citep{chao2025anomalyr1}, IAD-R1 \citep{li2025iadr1} and AnomalyThink \citep{acudad2026anomalythink}. In both categories the explanation itself is rarely evaluated. MMAD \citep{jiang2025mmad} measures reasoning as multiple-choice accuracy, VisA-D\&R \citep{xu2025anomalyov} uses sentence similarity and a GPT score, and SAGE \citep{zang2025sage} scores identification, localisation and consistency with a prompted GPT-4o rubric. None of these validates the evaluator against human ratings, and none reports its cost.

\paragraph{LLM Judges.}
Reference-based metrics such as BLEU, ROUGE and BERTScore
\citep{papineni2002bleu, lin2004rouge, zhang2020bertscore}
often agree poorly with human judgements of open-ended text,
leading to the adoption of LLM-as-a-judge evaluation
\citep{zheng2023judging, liu2023geval}.
LLM judges are costly, non-deterministic and can be biased toward
fluent prose \citep{zheng2023judging, huang2025empirical}.
For AD, \citet{ovad2026} find that a GPT-4o judge rewards
fluent but factually incorrect explanations.
For video anomaly understanding, MMEval \citep{du2024cuva}
and VAU-Eval \citep{zhu2025vaur1} use rubric-based judging,
while FV-Score \citep{pereira2026finevau} evaluates What, Who
and Where claims using an LLM.
Our approach shares this claim-level perspective
\citep{min2023factscore, zha2023alignscore}, but extracts claims
without token generation and verifies them deterministically
against ground truth rather than reference text.

\section{Proposed Method}

\subsection{Problem Definitions}
\noindent\textbf{Evaluation of Anomaly Explanation.}  Let \(x\) denote an inspection image with ground truth
\(g=(y,m,\mathcal{A})\), where \(y\) is the binary anomaly label,
\(m\) is the pixel-level anomaly mask, and \(\mathcal{A}\) denotes the
set of available ground-truth attributes associated with the image.
Given \(x\), a VLM produces a free-text explanation \(e\) describing its prediction and supporting evidence. Existing evaluation often considers only the predicted anomaly verdict, leaving most of the explanation unevaluated. Our goal is to assess whether the claims made in \(e\) are both faithful to and complete with respect to the available ground-truth information in \(g\). We formulate this as claim extraction followed by claim verification. \smallskip

\noindent\textbf{Claim Extraction.}
Our goal is to identify what \(e\) asserts, regardless of whether those
assertions are correct. Since explanations may take arbitrary forms, we map
\(e\) into a fixed set of structured claims,
\(c=f(e)=(\hat y,\{\hat a_q\}_{q=1}^{Q})\), where \(\hat y\) denotes the
predicted verdict expressed in the explanation and each \(\hat a_q\) denotes
the claim associated with a verifiable attribute \(q\). This representation
captures both stated and omitted information while remaining comparable across
explanations of different forms. Importantly, the extractor observes only the
explanation text and has no access to the image or ground truth \(g\). \smallskip

\noindent\textbf{Claim-based Verification.}
Once extracted, the claims are verified against their corresponding ground truth annotations and aggregated into an explanation level score \(S(e,g)\), without requiring a reference explanation.

\subsection{AnomalyJev}
\begin{figure*}
    \centering    
    \includegraphics[width=1\linewidth]{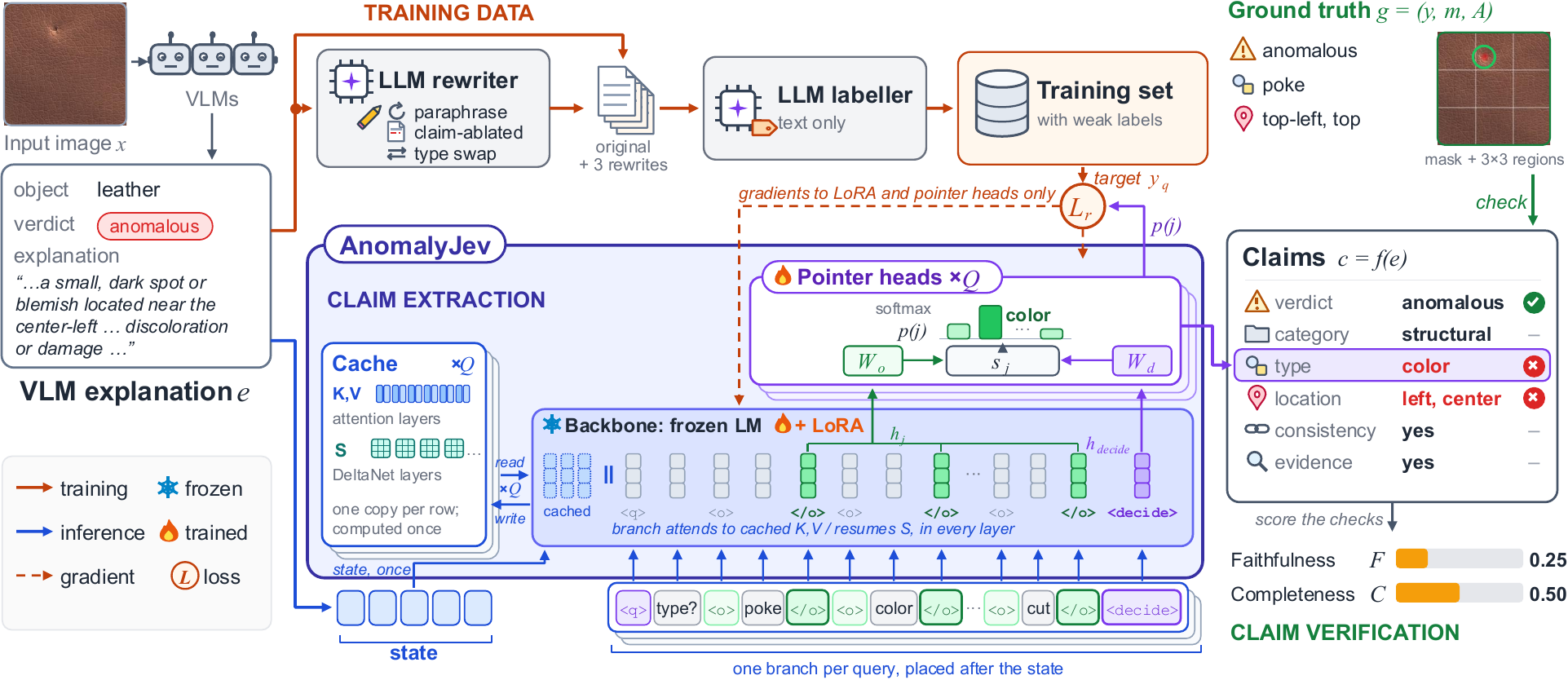}
    \caption{Overview of \sys{}.
\textbf{Training} (orange): VLM explanations are augmented and weakly labelled by an LLM to supervise the LoRA module and query specific pointer heads.
\textbf{Inference} (blue): the explanation is encoded once, question branches are processed in parallel, and the pointer head selects one option from each question's option set without token generation. The extracted claims are verified against the ground truth to score faithfulness and completeness.}
    \label{fig:framework}
\end{figure*}

\paragraph{Overview.} As shown in Fig.~\ref{fig:framework}, AnomalyJev evaluates the faithfulness and completeness of MLLM-generated anomaly explanations through a two-stage process grounded in the available ground truth. It first uses a lightweight architecture to extract verifiable claims from each explanation, then checks these claims against the ground truth to compute the respective scores for both criteria.

\paragraph{Verifiable Claim Extraction.} Since each verifiable attribute has a closed set of possible values, claim extraction can be formulated as selecting the option expressed in the explanation for each attribute. AnomalyJev performs this selection directly from contextualised token representations using a dedicated prediction head for each query, without requiring autoregressive decoding or an evaluation rubric at inference time.

To generate contextualised representations, we employ a small LLM to jointly encode the model-generated explanation and the information to be extracted. For each attribute \(a\), the extraction instruction is formulated as one or more queries \(q\), each with an associated option set \(\mathcal{O}_q\):
\begin{equation}
\begin{aligned}
u_q ={}& \texttt{<q>}~q \quad \texttt{<o>}o_1\texttt{</o>}~\cdots~
        \texttt{<o>}o_{|\mathcal{O}_q|}\texttt{</o>} \\
      & \texttt{<decide>},
\qquad o_i \in \mathcal{O}_q .
\end{aligned}
\end{equation}
where \texttt{<q>}, \texttt{<o>}/\texttt{</o>}, and \texttt{<decide>} denote the query token, the start/end option tokens, and the final decision token, respectively. The full set of extraction queries is listed in Table~\ref{tab:queries}. Because the queries are independent, they can be executed in parallel. We structure the input as a set of explanation–query pairs, with each pair targeting a single attribute. The explanation representation only needs to be computed once and can be cached, while all questions are batch encoded for efficiency.

To extract the claims for each attribute, the backbone encodes the explanation followed by the corresponding query, its candidate options, and a final \(\langle\texttt{decide}\rangle\) token.
Let \(h_j \in \mathbb{R}^{d_t}\) denote the hidden state of the closing token of option \(j\), and \(h_{\mathrm{decide}} \in \mathbb{R}^{d_t}\) that of \(\langle\texttt{decide}\rangle\).
Each option is scored by a scaled dot product,
\begin{equation}
s_j = \frac{(W_{\mathrm{o}} h_j + b_{\mathrm{o}})^\top (W_{\mathrm{d}} h_{\mathrm{decide}} + b_{\mathrm{d}})}{\sqrt{d_p}},
\end{equation}
where \(W_{\mathrm{o}}, W_{\mathrm{d}} \in \mathbb{R}^{d_p\times d_t}\) and \(b_{\mathrm{o}}, b_{\mathrm{d}} \in \mathbb{R}^{d_p}\) project the option and decision states into a \(d_p\)-dimensional space, and the answer distribution is
\begin{equation}
p(j) = \operatorname{softmax}_{j\in\mathcal{O}_q}(s_j).
\end{equation}
This can be viewed as a single-query attention scorer in the style of pointer networks~\cite{vinyals2015pointer}, where \(\langle\texttt{decide}\rangle\) attends over the candidate option representations and the resulting attention distribution directly defines the answer distribution. The prediction is therefore always constrained to the provided option set.

\paragraph{Scoring Metrics}
By verifying the extracted claims against the ground truth, we define two
scores that assess explanation quality in terms of \emph{faithfulness} and
\emph{completeness}. Let \(\mathcal{A}_x\) denote the set of attributes that
can be verified for image \(x\). In our setting,
\(\mathcal{A}_x=\{\text{type},\text{location}\}\) for anomalous images and \(\mathcal{A}_x=\emptyset\) for normal images. For each attribute \(a\in\mathcal{A}_x\), let \(\sigma_a=1\) if the explanation states \(a\), and \(c_a=1\) if the stated value matches the ground truth (for location: a claimed cell of a \(3\times3\) grid overlaps the mask). 
We further define
\[
n=\sum_{a\in\mathcal{A}_x}\sigma_a,\quad
w=\sum_{a\in\mathcal{A}_x}\sigma_a(1-c_a),\quad
K=|\mathcal{A}_x|,
\]
where \(n\) is the number of stated attributes and \(w\) is the number of
incorrectly stated attributes. Let \(v_{\mathrm{ok}}\in\{0,1\}\) indicate
whether the explanation's anomaly verdict is consistent with the ground
truth. \smallskip

\noindent\textit{Faithfulness.} Faithfulness measures whether the claims made by an explanation are correct.
We penalise the first incorrect claim by \(\beta\), while further errors
incur a proportional penalty controlled by \(\gamma\):
\begin{equation}
F(e,x)=v_{\mathrm{ok}}
\left[
1-\beta\,\mathbf{1}[w\geq1]
-\gamma\frac{(w-1)_+}{\max(n-1,1)}
\right]
\end{equation}
where \((z)_+=\max(z,0)\). Thus, an explanation that makes no incorrect
claim receives full faithfulness, while additional incorrect details further
reduce the score. If no attribute is stated, \(n=0\), faithfulness depends
only on the correctness of the overall verdict. For images with combined defects, the type cannot be checked against a single label and is not counted. 
\textit{Completeness.}
Completeness measures how much of the verifiable ground-truth information is correctly covered by the explanation. For normal images, $C(e,x)=v_{\mathrm{ok}}$.
For anomalous images, a fraction $(1-\alpha)$ of the score is assigned to the
correct anomaly verdict, while the remaining $\alpha$ is determined by
correctly stated attributes:
\begin{equation}
C(e,x)=
v_{\mathrm{ok}}
\left[
(1-\alpha)
+\frac{\alpha}{K}
\sum_{a\in\mathcal{A}_x}\sigma_a c_a
\right].
\end{equation}
Missing attributes therefore receive no credit, as do incorrectly stated
ones. Two cases earn no attribute credit even when stated: on images with
several combined defects, a single stated type covers only part of the
anomaly (\(c_{\text{type}}=0\)), and when the defect covers most of the image,
a stated location carries no information (\(c_{\text{loc}}=0\)).
Unlike faithfulness, which evaluates only the claims that are made,
completeness rewards explanations for covering more of the available
ground-truth.

\paragraph{Training AnomalyJev.}
The extraction module of AnomalyJev can be trained using weak supervision from LLM-generated labels, without requiring human annotation. We use a small set of VLMs, both anomaly-specific and general-purpose, to generate explanations for the training images, and retain both the original explanations and augmented variants produced through paraphrasing, defect-type swapping, and claim ablation. We then use an LLM to extract structured claims from each explanation by selecting from predefined options, using only the explanation text without access to the image ground truth.
We fine-tune the small LLM component using LoRA adapters, while
training the pointer-network extraction head from scratch.
For each question \(q\), we apply cross-entropy over only its candidate
option set \(\mathcal{O}_q\), where \(y_q\) denotes the labelled option.
For each explanation \(r\), we minimise the average cross-entropy over
questions \(Q_r\) with valid labels:
\begin{equation}
\mathcal{L}_r =
-\frac{1}{|Q_r|}
\sum_{q\in Q_r}
\log
\frac{\exp(s_{y_q})}
{\sum_{j\in\mathcal{O}_q}\exp(s_j)} ,
\end{equation}
where \(y_q\) is the labelled option and \(\mathcal{O}_q\) is the
candidate option set for question \(q\). The final training loss is the
mean of \(\mathcal{L}_r\) over the explanations in each optimisation
update.

The backbone is frozen except for LoRA adapters \citep{hu2021lora}, which are trained jointly with the pointer heads.
At inference, no decoding, prompting or output parsing is used, so the extractor is deterministic and its cost is one pass over the explanation plus one batched pass over the queries.

\section{Experiments}
Our experiments aim to answer three questions. (i) How well does \sys{} agree with human experts on faithfulness and completeness compared with existing evaluators, how complementary are the two metrics, and at what cost (Sec.~\ref{sec:human}, \ref{sec:comp} \ref{sec:efficiency})?
(ii) Do anomaly-specialised MLLMs produce better explanations than general-purpose ones (Sec. \ref{sec:benchmark})?
(iii) Does \sys{} generalise to unseen MLLMs and defect vocabularies (Sec.~\ref{sec:generalisation})?
We also examine key design choices through ablation studies (Sec.~\ref{sec:ablation}).

\subsection{Experimental Details}
\label{sec:exp_details}

\noindent\textbf{Implementation Details.}
\sys{} fine-tunes Qwen3.5-0.8B-Base with LoRA (rank 16, $\alpha=32$, dropout .05, all linear layers) and a pointer head of dimension 256. Training explanations are generated by AnomalyGPT, Qwen3.5-9B and Gemini 3.5 Flash on MVTec-AD and MVTec LOCO, and labelled by Gemini 3.5 Flash from the text alone. Gemini 3.5 Flash also writes the augmentations: a paraphrase of each explanation, a rewrite that keeps only the verdict and location, and, for 2,259 explanations, a rewrite that claims a different defect type. For half of the records, the defect-type options include distractor types from other objects. The final set has 22,521 records (6,886 original) and 337,255 questions. We train for one epoch with AdamW (learning rate $10^{-4}$, weight decay .01, effective batch size 8) in bf16, which takes 3.3 hours on one A100. At inference, each question takes the highest-scoring option. Scoring uses $\alpha=\beta=.5$ and $\gamma=.25$; locations are mapped to a $3\times3$ grid, and a mask covering at least 7 of the 9 cells gives no location credit. \smallskip

\noindent\textbf{Evaluator Baselines.}
We group baselines by the ground truth available to each evaluator (GT column in Table~\ref{tab:human}).
\emph{Conventional metrics} include BLEU-4, ROUGE-L and BERTScore, which compare explanations with ground-truth-derived references (\emph{Ref}), and CLIPScore and G-Eval~\citep{liu2023geval}, which use no ground truth. G-Eval uses a one-line 1--5 scoring prompt (\emph{S}).
\emph{VLM judges} receive the image with the ground-truth anomaly region highlighted and the true defect type (\emph{Ov}). For completeness, they also receive textual ground truth (\emph{W}), as provided to human raters.
\emph{Holistic} judges assess overall quality without specific criteria (\emph{Q}), while \emph{criteria-based} judges follow the human rater instructions (\emph{F}). We additionally test Qwen3-VL-30B with textual ground truth for faithfulness, both with and without the image.
\emph{Claims + Rule} replaces \sys{}'s extractor with Gemini 3.5 Flash while retaining the same verification rule (\emph{R}).
Methods producing a single score are compared against both human ratings. Further details are in Appendix~\ref{app:baselines}. \smallskip

\noindent\textbf{Evaluation Metrics.}
We assess how well each metric agrees with human experts on faithfulness and completeness at both the explanation and model levels. At the explanation level, we use Spearman's $\rho$ \cite{spearman1904proof} to measure the correlation between automatic scores and human ratings. At the model level, we use Kendall's $\tau$ \cite{kendall1938new} to compare the rankings of the 11 VLMs based on their mean scores. Both measures account for ties.

\begin{table*}[t]\centering
\begin{adjustbox}{max width=\textwidth}
\begin{tabular}{llccc cccc cccc}\toprule
 & & & & & \multicolumn{4}{c}{Faithfulness} & \multicolumn{4}{c}{Completeness}\\
\cmidrule(lr){6-9}\cmidrule(lr){10-13}
 & & & & & \multicolumn{2}{c}{R1} & \multicolumn{2}{c}{R2} & \multicolumn{2}{c}{R1} & \multicolumn{2}{c}{R2}\\
\cmidrule(lr){6-7}\cmidrule(lr){8-9}\cmidrule(lr){10-11}\cmidrule(lr){12-13}
Type & Method & Img & GT & Rub & $\rho$ & $\tau$ & $\rho$ & $\tau$ & $\rho$ & $\tau$ & $\rho$ & $\tau$\\\midrule
\multirow{5}{*}{Conventional} & BLEU-4 & -- & Ref & -- & .205 & $-$.127 & .197 & $-$.091 & .124 & $-$.150 & .214 & .170\\
 & ROUGE-L & -- & Ref & -- & .314 & $-$.200 & .307 & $-$.164 & .251 & .000 & .316 & .321\\
 & BERTScore & -- & Ref & -- & .245 & $-$.164 & .236 & $-$.200 & .239 & $-$.075 & .308 & .245\\
 & CLIPScore & \checkmark & -- & -- & .113 & .164 & .128 & .127 & .216 & .187 & .166 & $-$.019\\
 & G-Eval (Gemini 3.5 Flash) & \checkmark & -- & S & .524 & .550 & .530 & .587 & .575 & .736 & .572 & .534\\\midrule
\multirow{3}{*}{Holistic judge} & Qwen3-VL-30B-A3B & \checkmark & Ov+W & Q & .852 & .891 & .856 & .927 & .819 & .876 & .799 & .712\\
 & Gemini 3.1 Pro & \checkmark & Ov+W & Q & .911 & .927 & .915 & \textbf{.964} & .818 & .476 & .806 & \textbf{.789}\\
 & Claude Sonnet 5.5 & \checkmark & Ov+W & Q & .909 & .927 & .914 & \textbf{.964} & .850 & .711 & .820 & .774\\\midrule
\multirow{5}{*}{Criteria judge} & LLaVA-Critic-R1-7B & \checkmark & Ov & F & .691 & .818 & .695 & .855 & .591 & .571 & .604 & .385\\
 & Qwen3-VL-8B & \checkmark & Ov & F & .724 & .891 & .725 & .927 & .732 & .726 & .711 & .594\\
 & Qwen3-VL-30B-A3B & \checkmark & Ov & F & .864 & .927 & .869 & .891 & .741 & .827 & .715 & .718\\
 & Gemini 3.1 Pro & \checkmark & Ov & F & \textbf{.967} & .927 & \textbf{.966} & .891 & .884 & .800 & .832 & .673\\
 & Claude Sonnet 5.5 & \checkmark & Ov & F & \textbf{.967} & \textbf{.945} & .965 & .945 & \textbf{.914} & \textbf{.914} & \textbf{.848} & .673\\\midrule
\multirow{4}{*}{GT as data}  & Qwen3-VL-30B-A3B & \checkmark & Ov+W & F & .946 & .927 & .946 & .891 & .741 & .827 & .715 & .718\\
 & Qwen3-VL-30B-A3B, text only & -- & W & F & .941 & .908 & .927 & .871 & .761 & .849 & .746 & .648\\
 & Gemini 3.5 Flash claims + rule & -- & Rule & R & .942 & .745 & .938 & .782 & .883 & .800 & .818 & .712\\
 & \sys{} (0.8B) & -- & Rule & R & .940 & .796 & .935 & .796 & .890 & .762 & .826 & .673\\\bottomrule
\end{tabular}
\end{adjustbox}
\caption{Agreement with two human raters (R1, R2) on faithfulness and completeness of MVTec. Baselines and column abbreviations are described in Sec.~\ref{sec:exp_details}. Best value per column in bold.}
\label{tab:human}
\end{table*}

\subsection{Data Preparation}
\noindent\textbf{Source Datasets.}
We construct the explanation corpus from AD datasets with image-level labels, pixel-level masks and defect types. The test splits of MVTec-AD \cite{bergmann2019mvtec} (structural defects across 15 categories) and MVTec LOCO \cite{bergmann2022loco} (structural and logical anomalies across 5 product categories) are used to construct the training and
in-distribution test sets, totalling 3,293 images. For cross-dataset evaluation, we sample 500 test images from 12 object categories in VisA \cite{zou2022visa}, without using any VisA data or defect vocabulary in training. Full dataset details are provided in Appendix~\ref{app:src_data}. \smallskip

\noindent\textbf{Explanation generation}. We collect explanations from five dedicated VLMs for anomaly detection, AnomalyGPT \cite{gu2024anomalygpt}, AnomalyR1 \cite{chao2025anomalyr1} and three AnomalyThink models \cite{acudad2026anomalythink} (built on Qwen3-VL-8B, Qwen2.5-VL-7B and LLaVA-OneVision-7B), and six general-purpose VLMs, Qwen3.5-9B \cite{qwen2026qwen35}, Gemini 3.5 Flash \cite{google2026gemini35flash}, InternVL3-8B \cite{zhu2025internvl3}, LLaVA-OneVision-7B \cite{li2024llavaonevision}, Gemma-3-12B \cite{gemma2025gemma3} and Qwen2.5-VL-3B \cite{bai2025qwen25vl}. All answer the same prompt for every image, asking for a verdict, the defect type and its location. Refer to Appendix~\ref{app:exp_gen} for further details. \smallskip

\noindent\textbf{Human-annotated sets.} Two experts independently rate 440 explanations (40 test images $\times$ 11 VLMs) for faithfulness (1--5) and a subset of 220 (20 per VLM) for completeness (1--3), blind to the VLM. They agree closely on both faithfulness and completeness (Spearman $\rho$ .984 / .903, weighted $\kappa$ .973 / .889, exact agreement 91.6\% / 89.1\%). For cross-dataset evaluation, one expert rates VisA explanations from 5 VLMs with the same instructions. Annotation details are given in Appendix~\ref{sec:annotation_apd}.

\subsection{Agreement with Human Evaluators}
\label{sec:human}
\noindent\textbf{Faithfulness.} Tab.~\ref{tab:human} reports agreement with the two human raters. Conventional metrics show weak agreement with human experts, providing limited meaningful assessment. Incorporating ground-truth information substantially improves agreement. For example, providing Qwen3-VL-30B with textual ground truth increases its faithfulness agreement from .864 to .946, demonstrating the importance of grounding evaluation in anomaly-specific information. \sys{} performs comparably to the best open MLLM judge (.940 vs.\ .946) and approaches the flagship judges (.967).

\noindent\textbf{Completeness.} Flagship judges achieve the strongest agreement with human experts, with Claude Sonnet 5.5 reaching Spearman's $\rho$ of .914 and .848 for the two raters. \sys{} performs comparably to Gemini 3.1 Pro ($\rho$ .890 vs.\ .884 for R1 and .826 vs.\ .832 for R2), while substantially outperforming all open MLLM judges ($\rho$ .59--.76), including the larger Qwen3-VL-30B-A3B provided with textual ground truth. On logical anomalies, \sys{} achieves the highest completeness agreement of all methods, while on structural defects its faithfulness agreement remains below flagship judges (Appendix~\ref{app:category}).

\noindent\textbf{Effect of Verifiable Criteria.} Holistic judges reflect common practice, where an MLLM rates overall explanation quality without explicit evaluation criteria. Compared to criteria-based judging, Gemini 3.1 Pro and Claude Sonnet 5.5 show lower agreement with human experts on both faithfulness ($\rho$ drops by .06) and completeness (.07), falling below \sys{}. This suggests that a single overall score does not reliably capture either aspect. Only the weaker Qwen3-VL-30B gains from holistic judging on completeness, possibly due to its limited ability to follow the detailed rubric. 

\sys{} achieves this from the explanation text alone, with a much smaller model and without the image, validating that claim-based verification can effectively assess anomaly-related information in terms of both metrics.

\subsection{Complementarity of Faithfulness and Completeness}
\label{sec:comp}
Tab.~\ref{tab:cf_agreement} shows that faithfulness and completeness assess different aspects of explanation quality. On the 220 explanations rated for both, the two metrics correlate only moderately for both human raters ($\rho=.79$). For the 165 defective images with a correct verdict, the correlation is weaker ($\rho=.67$). This is mostly caused by explanations that are accurate but vague (e.g.\ ``an anomaly at the left of the image'') or that mix correct and incorrect details. \sys{}'s completeness score catches this case, although it separates the two aspects less  than the raters ($\rho=.89$ and $.76$). This matters for potential metric hacking, because faithfulness alone does not penalise trivial, overly simple answers. In the extreme case of verdict-only explanations, every MLLM would rank first or second on faithfulness but very differently on completeness.

\subsection{Do Anomaly MLLMs Explain Better?}
\label{sec:benchmark}
We evaluate all 11,586 explanations in the corpus using \sys{}. Overall, anomaly-specialised VLMs show no consistent advantage over general-purpose VLMs. Although they achieve slightly higher average scores for faithfulness (.503 vs.\ .439) and completeness (.543 vs.\ .514), neither difference is statistically significant (permutation test, $p=.37$ and $.42$). The general-purpose LLaVA-OneVision-7B achieves the highest completeness score, while the specialised AnomalyR1 ranks ninth. Per-model rankings by faithfulness and completeness are given in Appendix~\ref{app:vlm_rank}. AnomalyGPT ranks first on faithfulness, largely because its templated explanations rarely include incorrect verifiable details. However, it often claims anomalies that are not present, which falls outside the current verification scope of \sys{} (Sec.~\ref{sec:limitations}). Logical anomalies remain challenging for all models, with completeness scores ranging from .36 to .49 on MVTec LOCO, compared with .50 to .75 on MVTec-AD.

\begin{table}[t]
\centering
\begin{adjustbox}{max width=0.8\columnwidth}
\begin{tabular}{lccc}
\toprule
 & \multicolumn{2}{c}{C vs.\ F, per explanation $\rho$} & VLM ranking \\
\cmidrule(lr){2-3}
Rater & all & correct  & $\tau$ \\
\midrule
Human R1 & .793 & .670 & .510 \\
Human R2 & .792 & .671 & .885 \\
\sys{} & .892 & .762 & .698 \\
\bottomrule
\end{tabular}
\end{adjustbox}
\caption{Spearman $\rho$ between faithfulness and completeness within each rater.}
\label{tab:cf_agreement}
\end{table}

\begin{table}[t]\centering\small
\setlength{\tabcolsep}{2.8pt}
\begin{adjustbox}{max width=\columnwidth}
\begin{tabular}{lcccccc}\toprule
 & \multicolumn{3}{c}{Faithfulness $\rho$} & \multicolumn{3}{c}{Completeness $\rho$}\\
\cmidrule(lr){2-4}\cmidrule(lr){5-7}
 & \multicolumn{2}{c}{MVTec VLMs} & & \multicolumn{2}{c}{MVTec VLMs} & \\
\cmidrule(lr){2-3}\cmidrule(lr){5-6}
Method & seen & unseen & VisA & seen & unseen & VisA\\\midrule
\multicolumn{7}{l}{\emph{Conventional metric}}\\
G-Eval (Gemini 3.5 Flash) & .362 & .622 & .593 & .614 & .574 & .504\\\midrule
\multicolumn{7}{l}{\emph{Image judges}}\\
Gemini 3.1 Pro & \textbf{.956} & .976 & \textbf{.931} & .900 & .871 & .861\\
Claude Sonnet 5.5 & .947 & \textbf{.977} & .930 & \textbf{.924} & \textbf{.886} & \textbf{.880}\\
Qwen3-VL-30B-A3B & .929 & .851 & .775 & .822 & .706 & .796\\\midrule
\multicolumn{7}{l}{\emph{Claims + rule}}\\
Gemini 3.5 Flash reader & .916 & .952 & .847 & .903 & .858 & .818\\
\sys{} 0.8B & .908 & .953 & .871 & .905 & .865 & .798\\\bottomrule
\end{tabular}
\end{adjustbox}
\caption{Generalisation to unseen VLMs and an unseen dataset (VisA): Spearman $\rho$ with human experts.}
\label{tab:transfer}
\end{table}

\subsection{Generalisation}\label{sec:generalisation}
In practice, new MLLMs are released much more frequently than new AD datasets, so evaluators are more likely to be applied to unseen models than unseen datasets.
We test two distribution shifts (Tab.~\ref{tab:transfer}): explanations from eight MLLMs excluded from training and explanations from an unseen dataset, VisA. \sys{} generalises to unseen MLLMs without a drop in faithfulness agreement ($\rho$ .908 seen vs.\ .953 unseen), while completeness agreement drops by only .04, comparable to flagship judges (Gemini 3.1 Pro .029, Claude Sonnet .038).
On VisA, \sys{} outperforms the open Qwen3-VL-30B judge on faithfulness (.871 vs.\ .775) and achieves comparable completeness (.798 vs.\ .796), but performs below flagship judges (.93 on faithfulness and .86–.88 on completeness). This gap mainly arises from location verification, as VisA objects occupy smaller portions of the image and explanations more often describe locations relative to object parts rather than image coordinates, which our grid-based verification cannot capture (Sec.~\ref{sec:limitations}). 

\subsection{Ablation Study}
\label{sec:ablation}
\begin{table}[t]\centering\small
\setlength{\tabcolsep}{4pt}
\begin{adjustbox}{max width=\columnwidth}
\begin{tabular}{lcccc}\toprule
 & \multicolumn{2}{c}{MVTec} & \multicolumn{2}{c}{VisA (unseen)}\\
\cmidrule(lr){2-3}\cmidrule(lr){4-5}
Variant & Faith. & Compl. & Faith. & Compl.\\\midrule
\multicolumn{5}{l}{\emph{(a) Claims used in the score}}\\
Verdict only & .871 & .608 & .751 & .703\\
\quad + type & .939 & .751 & \textbf{.893} & \textbf{.773}\\
\quad + location & .881 & .801 & .779 & .720\\
All claims (\sys{}) & \textbf{.940} & \textbf{.877} & .871 & .759\\\midrule
\multicolumn{5}{l}{\emph{(b) Extraction head (no augmentation)}}\\
Linear probe & .934 & .778 & .698 & .663\\
Classification heads (full FT) & .936 & .863 & .850 & .725\\
Pointer head & \textbf{.940} & \textbf{.870} & \textbf{.860} & \textbf{.761}\\\midrule
\multicolumn{5}{l}{\emph{(c) Training augmentation (pointer head)}}\\
None & .940 & .870 & .860 & \textbf{.761}\\
Claim-ablated + type-swap rewrites & .940 & \textbf{.877} & \textbf{.871} & .759\\\midrule
\multicolumn{5}{l}{\emph{(d) Backbone size (with augmentation)}}\\
0.8B (\sys{}) & .940 & .877 & \textbf{.871} & .759\\
2B & .941 & .870 & .851 & .750\\
4B & \textbf{.943} & \textbf{.878} & .851 & \textbf{.775}\\\bottomrule
\end{tabular}
\end{adjustbox}
\caption{Ablations: Spearman $\rho$ with human experts. Each block ablates one factor of the default model.}
\label{tab:ablation}
\end{table}

We analyse the effectiveness of the key design choices in \sys{} and report their agreement with human judgements in Tab.~\ref{tab:ablation}. \smallskip

\noindent\textbf{Multiple Claims.}
We compare \sys{} with variants that use only the verdict, or the verdict with either the defect type or location.
On MVTec(in-domain), \sys{} achieves the strongest agreement with human experts, particularly for completeness ($\rho=.877$ vs.\ $.608$ for the verdict alone). For faithfulness, most of the improvement comes from including anomaly type. On VisA, removing the location claim has little effect on agreement, with the difference falling within the confidence interval. This may due to VisA explanations often describe defect locations relative to specific object parts, making location claims harder to extract reliably. \smallskip

\noindent\textbf{Pointer Head.} We replace the pointer head with classification heads trained through full fine-tuning or linear probing. All three variants achieve similar in-domain faithfulness agreement, but linear probe shows lower agreement on completeness.
The pointer head generalises best to unseen defect types on VisA, with a completeness correlation of .761 compared with .725 and .663. This suggests that selecting among the options in each attribute query transfers better to new defect vocabularies. \smallskip

\noindent\textbf{Augmentation.}~Augmentation leaves agreement unchanged; its benefit is robustness: without it, the extractor reads every detail-free answer (``Yes, there is a defect in this X'') as naming a defect type (0\% correct vs.\ 83\%; Appendix~\ref{app:robust}). \smallskip

\noindent\textbf{Backbone size.} Scaling to 4B does not improve agreement, so the remaining error lies in the scoring rule rather than the reader, and the 0.8B model suffices.
Results are stable under alternative scoring weights and generalise to the second expert (Appendix~\ref{app:rule}).

\subsection{Efficiency}
\label{sec:efficiency}
\begin{figure}[t]
\centering
\includegraphics[width=\columnwidth]{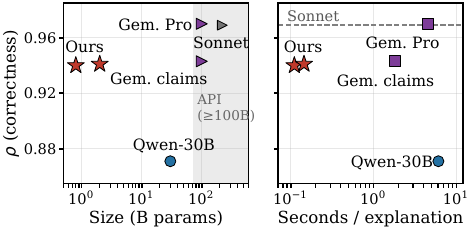}
\caption{Agreement with human experts (Spearman $\rho$, faithfulness) against model size and runtime.}
\label{fig:efficiency}
\end{figure}

Figure~\ref{fig:efficiency} compares agreement with human experts against model size and runtime. \sys{} scores an explanation in 0.11\,s on a single A100 (\$0.06 per 1,000 explanations), about 40$\times$ faster and 150$\times$ cheaper than the Gemini 3.1 Pro judge (4.56\,s, \$8.85), while its agreement is .03 lower ($\rho=.940$ vs.\ $.970$). Open VLM judges are both slower and less accurate. For example, Qwen3-VL-30B takes 6.06\,s per explanation and reaches $\rho=.871$. Scoring the full corpus of 11,586 explanations takes about 21 minutes with \sys{}, compared with roughly 15 hours and \$100 with Gemini 3.1 Pro.

\section{Conclusion}
In this paper, we propose claim-based verification for image anomaly explanations, grounding evaluation in anomaly-relevant information. We instantiate this idea in AnomalyJev, a lightweight framework that extracts verifiable claims using a small language model and checks them against the available ground truth to assess explanation faithfulness and completeness. We construct a human-annotated evaluation set of 640 explanations with over 1,600 expert ratings to validate the framework. Extensive experiments demonstrate that AnomalyJev achieves stronger agreement with human judgements than conventional metrics and most open MLLM judges, while approaching that of flagship proprietary judges. Its lightweight design and generation-free claim extraction further enable efficient and scalable evaluation at substantially lower computational cost.

\section{Limitation and Broader Impact}
\label{sec:limitations}
Although \sys{} achieves competitive agreement with human experts, its evaluation is limited to the faithfulness and completeness of information supported by the available ground truth. Claims that fall outside this ground truth cannot be reliably assessed. For example, AnomalyGPT often states the number of anomalies (``there are 2 anomalies''), which our scoring rule does not check, and such claims go unverified. The method is also less informative for datasets with only binary labels. Locations are verified on a $3\times3$ image grid, which cannot capture positions described relative to object parts; resolving such positions requires locating the parts in the image, which \sys{} avoids by design to remain lightweight.

Our human evaluation covers two experts on MVTec and one on VisA, with a few hundred explanations each. When applying \sys{} directly to a new domain, validation against human-annotated data is therefore recommended. In addition, when an application prioritises aspects beyond anomaly-relevant faithfulness and completeness, \sys{} may produce rankings that differ from evaluations tailored to those objectives. Finally, \sys{} scores explanations rather than inspection decisions; in safety-critical settings, high scores should not replace human quality assurance.

\bibliography{custom}
\newpage

\appendix

\section{Additional Experimental Details}
\subsection{Source Dataset Details}\label{app:src_data}
MVTec LOCO's logical anomalies include missing, extra or misplaced parts and wrong quantities. The VisA sample is stratified by object and label. For every image we record its label, anomaly category (structural, logical or none) and defect type, taking type names from the dataset annotations (73 in MVTec-AD, 89 in MVTec LOCO and 71 in VisA, counted per object category). Each image is resized to 448 pixels on the longer side before it is given to a VLM.

\subsection{Explanation generation.}\label{app:exp_gen}
\begin{table}[h]
\centering
\begin{adjustbox}{max width=\columnwidth}
\begin{tabular}{lcccc}
\toprule
MLLM & AD & LOCO & Total & VisA\\
\midrule
AnomalyGPT$^{\dagger\ast}$ & 1,725 & 1,568 & 3,293 & --\\
Qwen3.5-9B$^{\ast}$ & 1,725 & 1,568 & 3,293 & 500\\
Gemini 3.5 Flash$^{\ast}$ & 523 & 477 & 1,000 & 500\\
AnomalyR1$^\dagger$ & 264 & 236 & 500 & 40\\
AT-Qwen3-VL-8B$^\dagger$ & 264 & 236 & 500 & 40\\
AT-Qwen2.5-VL-7B$^\dagger$ & 264 & 236 & 500 & --\\
AT-LLaVA-OV-7B$^\dagger$ & 264 & 236 & 500 & 40\\
InternVL3-8B & 264 & 236 & 500 & --\\
LLaVA-OneVision-7B & 264 & 236 & 500 & --\\
Gemma-3-12B & 264 & 236 & 500 & --\\
Qwen2.5-VL-3B & 264 & 236 & 500 & --\\
\midrule
Total & 6,105 & 5,481 & 11,586 & 1,120\\
\bottomrule
\end{tabular}
\end{adjustbox}
\caption{Explanations per MLLM. AD: MVTec-AD; LOCO: MVTec LOCO; AT: AnomalyThink; $^\dagger$dedicated anomaly MLLM; $^\ast$explanations used for training \sys{}.}
\label{tab:corpus}
\end{table}
The three AnomalyThink models are built on Qwen3-VL-8B, Qwen2.5-VL-7B and LLaVA-OneVision-7B. Every VLM receives the prompt ``Is there a defect in this [object]? Answer yes/no, then describe the defect type and where it is.'' with greedy decoding and thinking disabled. AnomalyGPT and Qwen3.5-9B answer all 3,293 MVTec images, Gemini 3.5 Flash a stratified sample of 1,000, and the remaining eight VLMs a shared stratified subset of 500, so that all 11 VLMs are compared on the same images. This yields 11,586 explanations (Table~\ref{tab:corpus}). On VisA, Qwen3.5-9B and Gemini 3.5 Flash answer all 500 sampled images, and AnomalyR1 and the Qwen3-VL-8B and LLaVA-OneVision AnomalyThink models answer the human-rated subset.

\subsection{Baseline Details}\label{app:baselines}
All VLM judges are evaluated zero-shot; LLaVA-Critic-R1-7B is itself trained for general-purpose response evaluation. The ground truth in words states whether the product is defective and which cells of the $3\times3$ grid the anomaly region touches. Holistic judges receive the same inputs as the completeness raters (overlay, true defect type and ground truth in words) and are asked to rate the overall quality of the answer from 1 to 5. Rule-based methods use no image.

\subsection{Extraction Queries}\label{app:queries}
Each explanation is queried with the 15 attribute queries in Table~\ref{tab:queries}. Every query instructs the extractor to judge only what the explanation claims, not whether the claim is true. Binary queries choose between \emph{true} and \emph{false}.

\begin{table*}[h]
\centering
\begin{adjustbox}{max width=\textwidth}
\begin{tabular}{llp{8.2cm}l}
\toprule
Query & Options & Instruction (abridged) & Used in \\
\midrule
Anomaly present & binary & Does the explanation claim that the object has an anomaly? A hedged claim counts. & $F$, $C$ \\
Defect type & object vocabulary + other / unspecified / none & Which defect type does the explanation claim for this [object]? Choose the listed type that matches in meaning; if several are claimed, choose the first (or \emph{combined}). & $F$, $C$ \\
Location: 9 regions & binary (each) & Does the explanation place a claimed defect in [region] of the $3\times3$ grid? Use only positional words; never infer a position from a part name. & $F$, $C$ \\
Location stated & binary & Does the explanation say where the claimed defect is, using positional language? & $F$, $C$ \\
Anomaly category & structural / logical / none / unspecified & What kind of anomaly does the explanation claim (local physical flaw vs.\ rule violation such as missing or misplaced parts)? & -- \\
Verdict consistent & binary & Does the explanation agree with the model's stated verdict? & -- \\
Cites evidence & binary & Does the explanation point to a concrete visual feature rather than asserting generically? & -- \\
\bottomrule
\end{tabular}
\end{adjustbox}
\caption{Attribute queries used by \sys{}. Defect-type options are object-specific (73 in MVTec-AD, 89 in MVTec LOCO, 71 in VisA).}
\label{tab:queries}
\end{table*}

\section{Human Annotation Details}
\label{sec:annotation_apd}
\subsection{Human Rater Details}
Our two human annotators are researchers with extensive experience in anomaly detection. Both are authors of this paper and received no additional compensation. The rated items are model-generated explanations of publicly available industrial inspection images. No personal or sensitive data is involved, so no ethics review was required.

\subsection{Human Labelling Interface}
\begin{figure*}[t]
    \centering
    \rotatebox{90}{\includegraphics[width=\textheight,height=\textwidth,keepaspectratio]{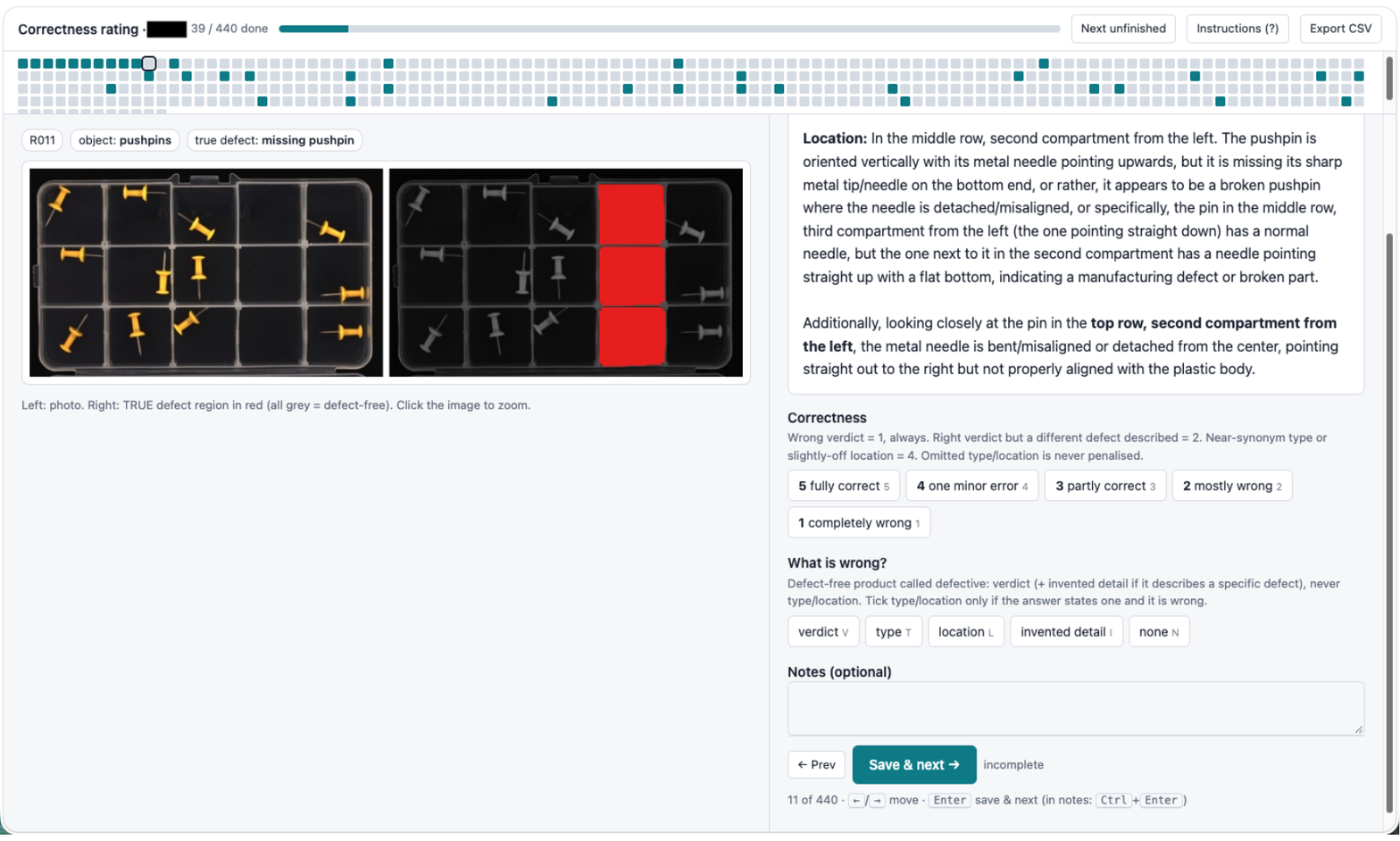}}
    \caption{Screenshot of the interface used by human raters.}
    \label{fig:interface}
\end{figure*}

Figure~\ref{fig:interface} shows the web interface used by the two experts. For each item, the image, ground-truth mask overlay and defect type are displayed on the left, with the explanation on the right and the generating MLLM hidden. Raters assign a faithfulness score from 1 to 5, identify incorrect claims (verdict, type, location or invented details), and can leave optional comments. Completeness is rated separately using textual ground truth. Items are presented in random order, and progress is saved after each rating.

\subsection{Human Annotation Guideline}
\label{app:guidelines}
The 40 images are stratified by dataset, label and object, among images for which all 11 VLMs produced explanations. Faithfulness ranges from 1 (wrong verdict) to 5 (verdict, type and location all correct); raters mark which parts are wrong, and omitted details are not penalised. Completeness ranges from 1 (no correct detail) to 3 (correct type and location). Raters see the image, the ground-truth mask overlay, the true defect type and, for completeness, the ground truth in words; VLM identity is hidden and rows are shuffled. Exact agreement between the two experts is 91.6\% (faithfulness) and 89.1\% (completeness). The VisA set covers 40 images.

\begin{figure*}[t]
\centering
\includegraphics[width=\textwidth]{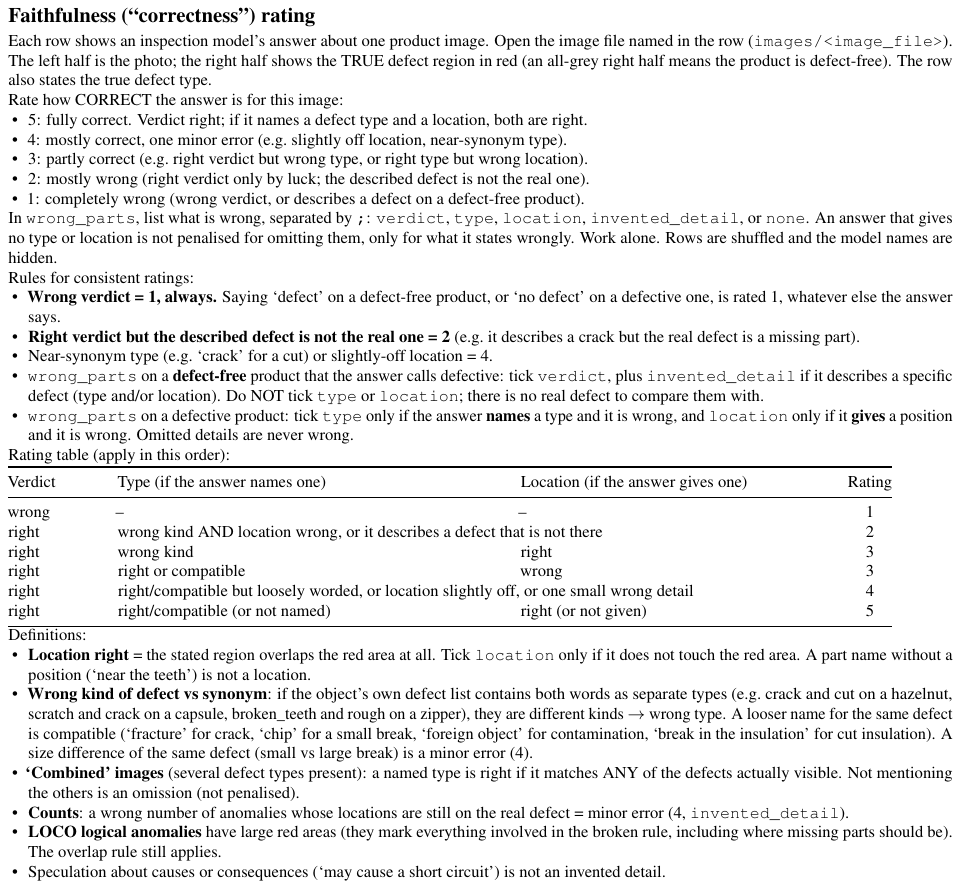}
\caption{Instructions given to the two experts for the faithfulness (``correctness'') rating.}
\label{fig:instr-faith}
\end{figure*}

\begin{figure*}[t]
\centering
\includegraphics[width=\textwidth]{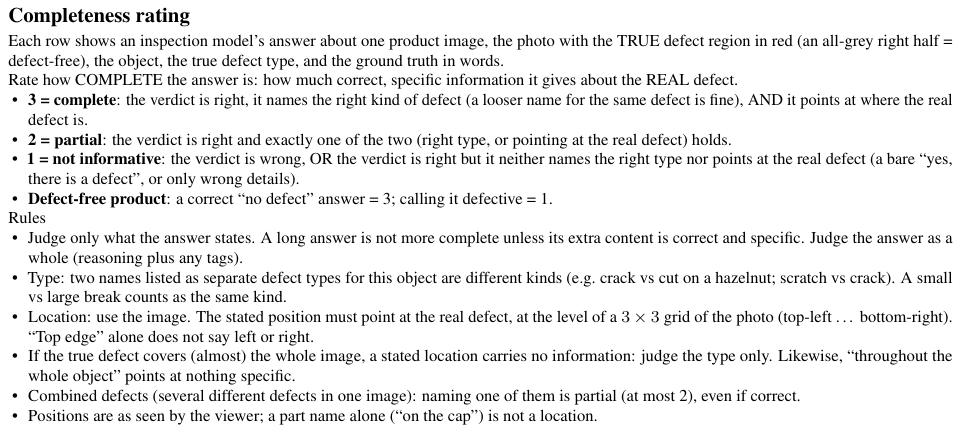}
\caption{Instructions given to the two experts for the completeness rating.}
\label{fig:instr-compl}
\end{figure*}

\paragraph{Full instructions shown to raters.} Faithfulness was rated under the label ``correctness''. Both instruction sets are reproduced verbatim in Figures~\ref{fig:instr-faith} and~\ref{fig:instr-compl}.

\section{Additional Experimental Results}

\subsection{Agreement by Anomaly Category}\label{app:category}
\begin{table}[h]
\centering
\begin{adjustbox}{max width=\columnwidth}
\begin{tabular}{lcccccc}
\toprule
 & \multicolumn{3}{c}{Faithfulness} & \multicolumn{3}{c}{Completeness}\\
\cmidrule(lr){2-4}\cmidrule(lr){5-7}
 & Struct. & Logical & Normal & Struct. & Logical & Normal\\
\midrule
$n$ & 154 & 66 & 220 & 132 & 43 & 45\\
\midrule
Gemini 3.1 Pro & \textbf{.842} & \textbf{.877} & .988 & .821 & .327 & 1.000\\
Claude Sonnet 5.5 & .839 & .874 & \textbf{.995} & \textbf{.832} & .686 & 1.000\\
\sys{} & .715 & .845 & \textbf{.995} & .804 & \textbf{.733} & 1.000\\
\bottomrule
\end{tabular}
\end{adjustbox}
\caption{Spearman $\rho$ with the mean of the two experts, by anomaly category.}
\label{tab:category}
\end{table}
Table~\ref{tab:category} splits agreement by anomaly category. Defect-free images are scored almost perfectly by all methods, as the verdict alone determines their rating; differences between methods therefore lie in defective images. On logical anomalies (MVTec LOCO), \sys{} agrees best with human experts on completeness. On structural defects, its faithfulness agreement is lower than that of flagship judges.

\subsection{VLM Rank}\label{app:vlm_rank}
Table~\ref{tab:vlm_rank} gives the full ranking of the 11 MLLMs by faithfulness and completeness.
\begin{table}[t]
\centering
\begin{adjustbox}{max width=0.8\columnwidth}
\begin{tabular}{lcccccc}
\toprule
 & \multicolumn{3}{c}{Faithfulness} & \multicolumn{3}{c}{Completeness}\\
\cmidrule(lr){2-4}\cmidrule(lr){5-7}
VLM & H1 & H2 & AJ & H1 & H2 & AJ\\
\midrule
LLaVA-OV-7B               & 1  & 1  & 1   & 3    & 3   & 3\\
AT-LLaVA-OV$^\dagger$     & 2  & 2  & 3.5 & 1    & 1   & 1\\
AT-Qwen3-VL$^\dagger$     & 3  & 3  & 3.5 & 5.5  & 8   & 7\\
AT-Qwen2.5-VL$^\dagger$   & 4  & 4  & 5   & 5.5  & 10  & 5.5\\
AnomalyGPT$^\dagger$      & 5  & 5  & 2   & 4    & 4   & 4\\
Gemini 3.5 Flash          & 6  & 6  & 6   & 2    & 2   & 2\\
InternVL3-8B              & 7  & 8  & 8.5 & 8.5  & 8   & 9\\
Qwen3.5-9B                & 8  & 7  & 8.5 & 7    & 5.5 & 10.5\\
AnomalyR1$^\dagger$       & 9  & 9  & 7   & 8.5  & 5.5 & 5.5\\
Qwen2.5-VL-3B             & 10 & 10 & 10  & 10.5 & 8   & 8\\
Gemma-3-12B               & 11 & 11 & 11  & 10.5 & 11  & 10.5\\
\midrule
$\tau$ with F rank        &    &    &     & .67  & .43 & .59\\
\bottomrule
\end{tabular}
\end{adjustbox}
\caption{Ranking of the 11 MLLMs by faithfulness and completeness on the human-rated explanations. H1/H2: human experts; AJ: \sys{}; AT: AnomalyThink; $^\dagger$dedicated anomaly MLLM.}
\label{tab:vlm_rank}
\end{table}

\subsection{Robustness to Perturbations}\label{app:robust}
We test whether claim extraction reacts to the content of an explanation rather than its wording, using 150 explanations and four perturbations:
\emph{paraphrase} (same claims, different wording; the extracted claims should not change),
\emph{type swap} and \emph{location swap} (a rewrite that claims a different defect type or location; the extracted claim should change accordingly), and
\emph{generic template} (``Yes, there is a defect in this X''; no type or location should be extracted).
Table~\ref{tab:robust} reports the results. Without augmentation, the extractor invents a defect type for every detail-free answer, because every training explanation that claims a defect also names one; adding claim-ablated rewrites to training fixes this.

\begin{table}[h]
\centering
\begin{adjustbox}{max width=\columnwidth}
\begin{tabular}{lcccc}
\toprule
Extractor & Paraphrase & Type swap & Location swap & Generic \\
\midrule
\sys{} & .927 & 1.000 & .939 & .827\\
\quad w/o augmentation & .853 & .880 & .901 & .000\\
Full fine-tuning & .807 & .200 & .847 & .413\\
Linear probe & .587 & .187 & .573 & .240\\
\bottomrule
\end{tabular}
\end{adjustbox}
\caption{Robustness of claim extraction (accuracy).}
\label{tab:robust}
\end{table}

\section{Scoring-Rule Sensitivity}\label{app:rule}
\begin{table}[h]
\centering
\begin{adjustbox}{max width=\columnwidth}
\begin{tabular}{llcc}
\toprule
 & Variant & MVTec & VisA\\
\midrule
\multirow{5}{*}{Faithfulness} & 1 / .5 / .25 (default) & .940 & .871\\
 & 1 / .67 / .33 & .940 & .871\\
 & 1 / .75 / .5 & .940 & .871\\
 & 1 / .25 / 0 & .901 & .774\\
 & 1 / 0 / 0 & .773 & .745\\
\midrule
\multirow{5}{*}{Completeness} & type .25 / loc.\ .25 (default) & .877 & .759\\
 & type .35 / loc.\ .15 & .872 & .759\\
 & type .15 / loc.\ .35 & .868 & .758\\
 & type only & .747 & .768\\
 & location only & .768 & .669\\
\bottomrule
\end{tabular}
\end{adjustbox}
\caption{Agreement (Spearman $\rho$) under alternative scoring weights.}
\label{tab:rule}
\end{table}

For faithfulness, the default rule assigns 1, .5 and .25 to answers with a correct verdict and zero, one or two wrong details. Spearman $\rho$ is unchanged under any order-preserving choice of these values; only merging levels lowers agreement. For completeness, the default weights type and location equally (.25 each); unequal weights change agreement little, while dropping either attribute lowers it. Table~\ref{tab:rule} reports $\rho$ with the mean of the two experts on MVTec and with one expert on VisA (100 explanations).

\end{document}